\documentclass[10pt,twocolumn,letterpaper]{article}
\usepackage{cvpr} 
\usepackage{graphicx}
\usepackage{amsmath}
\usepackage{amssymb}
\usepackage{booktabs}

 \usepackage{float} 

\usepackage[pagebackref,breaklinks,colorlinks]{hyperref}

\usepackage[capitalize]{cleveref}
\crefname{section}{Sec.}{Secs.}
\Crefname{section}{Section}{Sections}
\Crefname{table}{Table}{Tables}
\crefname{table}{Tab.}{Tabs.}

\usepackage{lipsum}
\usepackage{xcolor}
\usepackage{makecell}
\usepackage{bbding}

\AtBeginDocument{%
  }

\begin{document}

\def\name{Media2Face: Co-speech Facial Animation Generation With Multi-Modality Guidance}

\title{\name}

\author{Qingcheng Zhao$^{1,2}$\thanks{Equal contribution.}
\and
Pengyu Long$^{1,2*}$
\and
Qixuan Zhang$^{1,2}$\thanks{Project Leader.}
\and
Dafei Qin$^{2,3}$
\and
Han Liang$^{1}$
\and
Longwen Zhang$^{1,2}$
\and
Yingliang Zhang$^{4}$
\and
Jingyi Yu$^{1}$\thanks{Corresponding author.}
\and
Lan Xu$^{1\ddag}$
\and \and
$^{1}$ShanghaiTech University \and $^{2}$Deemos Technology \and $^{3}$University of Hong Kong \and $^{4}$DGene Digital Technology Co., Ltd.
\and
{\tt\small \{zhaoqch1, longpy, zhangqx1, lianghan, zhanglw2, yujingyi, xulan1\}@shanghaitech.edu.cn} \and {\tt\small qindafei@connect.hku.hk} \and {\tt\small yingliang.zhang@dgene.com}
\and
\small\url{https://sites.google.com/view/media2face}
}

\twocolumn[{%
\renewcommand\twocolumn[1][]{#1}%
\maketitle
\vspace{-30pt}
\begin{center}
    \centering
    \captionsetup{type=figure}
    \includegraphics[width=\textwidth]{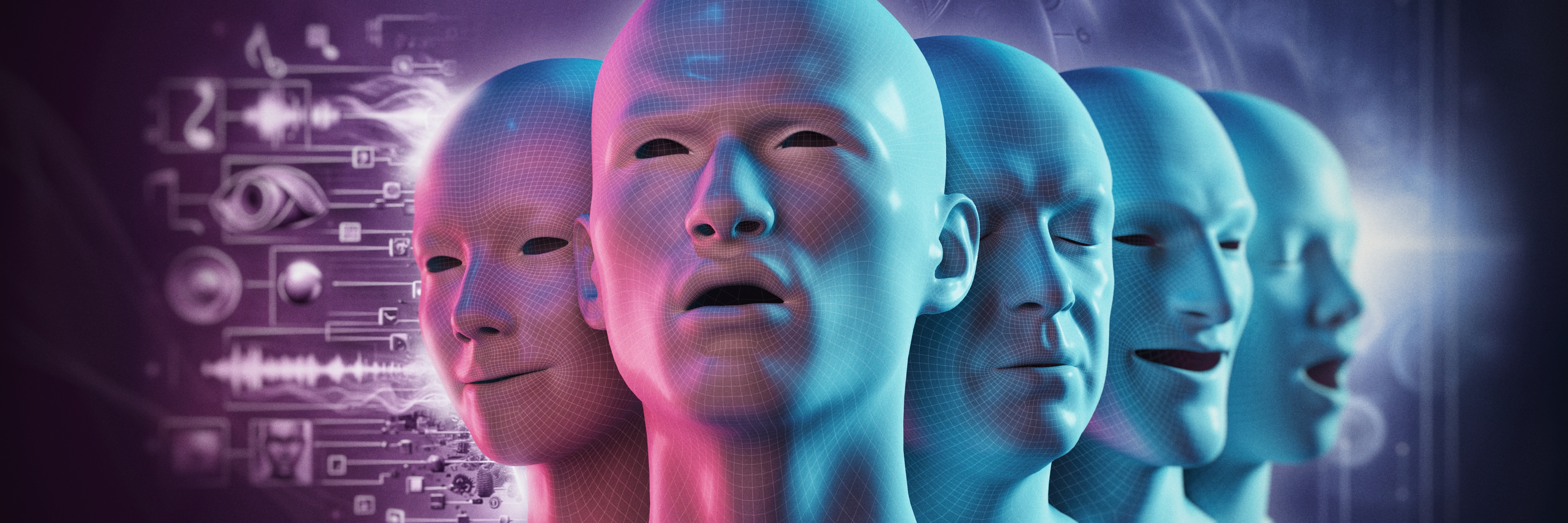}
    \captionof{figure}{Given the speech signal and multi-modal conditions (\textit{Left}), 
our system generates personalized and stylized co-speech facial animation and head poses (\textit{Middle, Right}).}
\end{center}%
}]

\begin{abstract}

The synthesis of 3D facial animations from speech has garnered considerable attention. 
Due to the scarcity of high-quality 4D facial data and well-annotated abundant multi-modality labels, 
previous methods often suffer from limited realism and a lack of flexible conditioning.
We address this challenge through a trilogy. 
We first introduce Generalized Neural Parametric Facial Asset (GNPFA), an efficient variational auto-encoder mapping facial geometry and images to a highly generalized expression latent space, decoupling expressions and identities. 
Then, we utilize GNPFA to extract high-quality expressions and accurate head poses from a large array of videos. This presents the M2F-D dataset, a large, diverse, and scan-level co-speech 3D facial animation dataset with well-annotated emotional and style labels.  
Finally, we propose Media2Face, a diffusion model in GNPFA latent space for co-speech facial animation generation, accepting rich multi-modality guidances from audio, text, and image. 
Extensive experiments demonstrate that our model not only achieves high fidelity in facial animation synthesis but also broadens the scope of expressiveness and style adaptability in 3D facial animation.
\end{abstract}

\section{Introduction}
\label{sec:intro}

Advancements in generative AI, powered by large language models, have brought to life virtual companion AI systems, reminiscent of the film ``Her''. The core of these systems is to provide realistic and immersive experiences, especially sustained emotional connections with users. 
To achieve this goal, it is crucial to generate natural facial animation consistent with rich speech content, subtle voice tones, and complicated underlying emotions. 

It has been a long journey for our graphics community to generate realistic co-speech facial animation. Early deterministic methods~\cite{edwards2016jali, MikeTalk,ekman1978facial} could generate only limited animation variations from audio. 
Recently, generative models, especially diffusion models, found their way to non-deterministic generation ranging from 2D image~\cite{pmlr-v37-sohl-dickstein15,ho2020denoising, salimans2022progressive} to 3D human motion~\cite{alexanderson2023listen,tevet2022human, liang2023omg, ao2023gesturediffuclip}.
Inspired by them, diffusion-based facial animation generation~\cite{thambiraja20233diface,aneja2023facetalk, chen2023diffusiontalker, sun2023diffposetalk, FaceDiffuser_Stan_MIG2023} has achieved promising results and hence received substantial attention. 

We observe two key factors to explore such diffusion-based generation schemes for facial animations.
First, the capability of the diffusion models heavily relies on large-scale and high-quality training data. 
However, most of the existing methods~\cite{xing2023codetalker,faceformer2022,FaceDiffuser_Stan_MIG2023} are trained on small-scale datasets such as VOCASET~\cite{Cudeiro_2019_CVPR} or BIWI~\cite{eth_biwi_00760}. These datasets only cover limited speech states and lack the diversity of emotional variations and character traits. 
Some recent methods~\cite{sun2023diffposetalk, EMOTE, yi2023generating, peng2023emotalk, 10.1007/978-3-031-20071-7_36} attempt to enrich speaking styles with 2D datasets. Yet, they fall short in authentically replicating natural expressions, since the adopted video-based facial trackers often produce sub-optimal expressions or neglect head motions. 
Second, it is crucial to enable flexible conditioning and disentangled controls, from diverse modalities like speech, style, or emotion.
Some concurrent methods offer keyframe~\cite{thambiraja20233diface} (3DiFace), implicit style~\cite{peng2023emotalk, sun2023diffposetalk}, or emoji-based~\cite{EMOTE} controls. Yet, faithful conditioning from more diverse multi-modalities like text and image inputs remains an open challenge.

In this paper, we approach the above challenges through a trilogy. First, we introduce General Neural Parametric Facial Asset (GNPFA), a neural representation of fine-grained facial expressions and head poses in latent space. We train GNPFA on a wide array of multi-identity 4D facial scanning data, including high-resolution images and artists' refined face geometries, dubbed Range of Motion (RoM) data. As such, we decouple nuanced facial expressions from identity in a latent representation that is generalizable to various talking styles.
Then, we utilize GNPFA to extract high-quality facial expressions and head poses from a diverse range of videos, including different content, styles, emotions, and languages. This results in the creation of the Media2Face Dataset (M2F-D), a diverse 4D dataset annotated with a variety of emotions and styles, with quality comparable to face scans.

Finally, we propose \textit{Media2Face}, a latent diffusion model for co-speech facial animation generation using the M2F-D dataset. It generates high-quality lip-sync with speech and expresses nuanced human emotions contained in text, images, and even music. Specifically, we train Media2Face in the latent space of GNPFA to recover fine-grained facial animations. It takes both audio features extracted by Wav2Vec2 ~\cite{baevski2020wav2vec} and text/image prompts encoded by CLIP ~\cite{radford2021learning} as conditions and generates the sequential facial expressions and head poses in a multi-classifier-free guidance manner.
We conduct extensive experiments and user studies to demonstrate the effectiveness of Media2Face. We further showcase various applications, i.e., generating vivid and realistic facial animations from diverse audio sources like dialogues, music, and speeches, as well as various text/image-based conditioning and editing. 
To summarize, our main contributions include: 
\begin{itemize} 

\item We present Media2Face, a diffusion-based generator that integrates diverse media inputs (audio, image, and text) to drive vivid facial animations including head poses.

\item To train Media2Face, We propose GNPFA, a neural latent representation to capture nuanced facial motion details, enabling the collection of a diverse co-speech 4D facial animation dataset with annotated expressions and styles.

\item We conduct extensive experiments and user studies and demonstrate exciting applications for generating facial animations with multi-modality guidances.

\end{itemize} 

\section{Related Work}
\label{sec:related}

\subsection{3D Facial Animation Representations}

The quality of generated 3D facial expressions heavily relies on the chosen representation method for 3D facial animation. Over the years, various representation methods have been proposed in the field~\cite{ekman1978facial, lewis2014practice, 10.1145/311535.311556}. One widely adopted approach is the Facial Action Coding System (FACS)~\cite{ekman1978facial}, which defines facial movements as combinations of muscle activations based on expert knowledge of human anatomy. Traditionally, these activations have been implemented using blendshape deformers~\cite{lewis2014practice}. To automate the process and encompass a broader range of facial expressions, \cite{10.1145/311535.311556} proposes 3D Morphable Models (3DMM). These models capture a linear deformation space directly from face scans, encompassing diverse identities and expressions~\cite{booth20173d, choe2006analysis, huber2016multiresolution}. For a comprehensive survey of 3DMM-related methods, we refer readers to \cite{egger20203d}. Despite their usefulness, linear methods have limitations in capturing subtle nuances of facial expressions. To address this, FLAME~\cite{10.1145/3130800.3130813} introduces pose-dependent corrective blendshapes, as well as articulated jaw, neck, and eyeballs, to enhance the fidelity of facial animation modeling. Further approaches have been developed to model facial expressions in nonlinear spaces~\cite{10.1016/j.patcog.2017.09.006, 10.2312:pgs.20141262}. ~\cite{10.1016/j.patcog.2017.09.006} employ a Gaussian mixture model to extend the traditional 3DMM, and~\cite{10.2312:pgs.20141262} introduce the use of multi-scale maps to transfer expression wrinkles from a high-resolution example face model to a target model.

Recent work~\cite{tran2018nonlinear, Tran_2019,tran2019highfidelity, Bouritsas_2019, chen2021learning, li2020learning, qin2023neural} leverages deep neural networks to build the latent expression space from data, achieving state-of-the-art performance on facial animation tasks. ~\cite{Bouritsas_2019} incorporates a specialized convolutional operator designed for 3D meshes, capitalizing on the consistent graph structure inherent to deformable shapes with unchanging topology. ~\cite{qin2023neural} learn interpretable and editable latent code over high-fidelity facial deformations, extending its application to various mesh topologies. These methods, however, are constrained by the scope of available datasets, often necessitating a compromise between enhancement of quality and retention of diversity. With the advancements in implicit representations, several work~\cite{yenamandra2020i3dmm,zheng2022imface,giebenhain2023nphm} have spurred efforts to employ Signed Distance Fields (SDFs) for modeling human head geometry, coupled with a forward deformation field to articulate facial expressions. However, they fall short in comparison to our approach, which constructs a latent space from a head mesh with a fixed topology. Our technique not only aligns better with practical applications but also ensures compatibility with traditional computer graphics (CG) pipelines.

\subsection{Conditional Facial Animation Synthesis}

\textit{Audio-Driven Facial Animation.} Recent studies in audio-driven 3D facial animation have leveraged procedural~\cite{MikeTalk, kalberer2001face} and learning-based methods~\cite{10.1145/1095878.1095881, 10.1145/3197517.3201292, 10.1145/3267935.3267950, 10.1145/3522615, richard2021meshtalk,faceformer2022, haque2023facexhubert, 10.1007/978-3-030-58517-4_42, 10.1145/3383652.3423911, chu2023corrtalk} to map speech to facial movements. While procedural techniques give artists control over lip shapes, they lack flexibility in capturing individual speech styles and require intensive manual tuning~\cite{edwards2016jali}. Learning-based approaches, using blendshapes~\cite{10.1145/3072959.3073699, https://doi.org/10.1111/cgf.14640, 10.1145/3242969.3243017, 9557828, peng2023emotalk} or 3D meshes~\cite{faceformer2022, Cudeiro_2019_CVPR, peng2023selftalk}, better capture speech nuances but can lead to overly smooth lip motion and limited upper face expressiveness. To address the complex speech-to-facial-expression mapping, probabilistic models like Vector Quantized Variational Autoencoders (VQ-VAE)~\cite{oord2018neural} have been introduced, predicting facial expression distributions from speech~\cite{Ng_2022_CVPR, xing2023codetalker, Yi_2023_CVPR}. Despite their strengths, these models often fail to fully represent the stochastic nature of facial expressions. Diffusion models~\cite{pmlr-v37-sohl-dickstein15, ho2020denoising}, recognized for handling intricate generative tasks~\cite{gal2022image, salimans2022progressive, Rombach_2022_CVPR, 10.1145/3626235}, show promise in audio-driven facial animation~\cite{sun2023diffposetalk, FaceDiffuser_Stan_MIG2023, chen2023diffusiontalker, thambiraja20233diface, aneja2023facetalk}, facilitating multi-modal generation. Yet, integrating speech with head movements remains a challenge, often yielding less convincing expressions. Our method introduces a novel head motion component within a prompt-guided diffusion model, improving the cohesiveness and expressiveness of generated facial animations, thus advancing the state-of-the-art in speech-driven 3D facial animation.

\begin{figure*}[thbp] 
  \includegraphics[width=\textwidth]{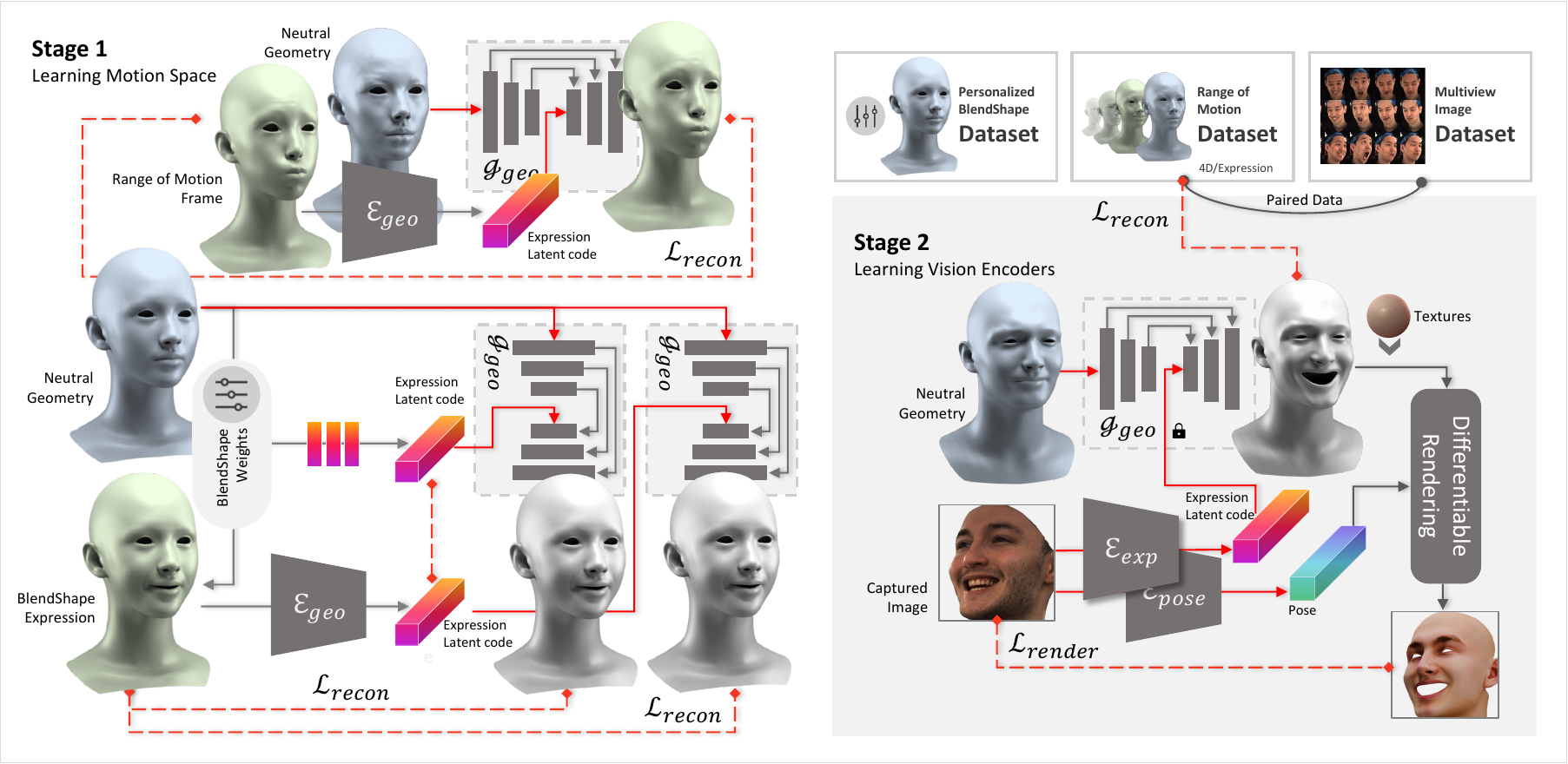}
  \caption{\textbf{GNPFA pipeline}. (\textit{Left:}) We train a geometry VAE to learn a latent space of expression and head pose, disentangling expression with identity. (\textit{Right:}) Two vision encoders are trained to extract expression latent codes and head poses from RGB images, which enables us to capture a wide array of 4D data.}
  \label{fig:NPFA}
\end{figure*}

\textit{Style Control.} Current approaches predominantly utilize two methods for style modulation: label-based and contextual-based control. Label-based control employs a fixed set of categories within the data to construct labels, such as emotion ~\cite{10.1145/3072959.3073658,EMOTE}, speaker style ~\cite{Cudeiro_2019_CVPR, faceformer2022}, and emotional intensity ~\cite{peng2023emotalk}. However, the pre-defined categories constrain the model to capture complex, nuanced emotions. Contextual-based control allows for the imitation of styles from given segments through a one-shot approach\cite{sun2023diffposetalk} or learning new speaking styles from given motions\cite{Thambiraja_2023_ICCV}. Techniques like 3DiFace\cite{thambiraja20233diface} adapt to the context of provided keyframes, while DiffPoseTalk\cite{sun2023diffposetalk} and Imitator\cite{Thambiraja_2023_ICCV} extract or learn style embeddings from given facial animations and style adaptations, respectively. Yet, these methods fall short of quality and explicit controls, often resulting in animations that lack the depth and complexity of genuine human expressions.

\section{Reshape Facial Animation Data}
\label{sec:data}

Realistic synthesis of 3D facial animations necessitates 4D dynamic facial performance capture, typically reliant on multiview-camera setups ~\cite{wuu2022multiface}. This requirement significantly limits the diversity and scalability of data acquisition due to the labor-intensive nature of capturing and processing such data.

To address these constraints, we propose Generalized Neural Parametric
Facial Asset (GNPFA), which is in essence a Variational Auto Encoder,  mapping facial geometry and video footprints to the same latent space.  We train GNPFA on large-scale 4D facial scanning, including high-resolution images and artists' refined geometries, enabling it to produce nuanced facial animation from videos with diverse identities, languages, emotions, and head poses.

\subsection{Expression Latent Space Learning}

\paragraph{Training data}
We first capture a dataset with vast multi-identity 4D scanings. We call this the Range of Motion (RoM) data. RoM consists of 43652 registered meshes and 698432 images from 300 identities across different genders, ages, and ethnicities. In addition, to enhance the robustness of expression-identity disentanglement, we create personalized blendshapes for 200 identities under FACS standards according to ~\cite{li2020dynamic}, and generate random artificial expressions during training for augmentation.

\paragraph{Geometry VAE}
To learn an expression latent space disentangled from identities, we design a geometry VAE consisting of a geometry encoder $\mathcal{E}_\text{geo}$ and a geometry generator $\mathcal{G}_\text{geo}$, where $\mathcal{G}_\text{geo}$ is conditioned on a neutral geometry and utilizes a UNet architecture. To support traditional blendshape animation, we train two mapping networks $\mathcal{M}$ and $\mathcal{M}'$, where the former maps the weight of blendshapes, $w$ to our latent space, and the latter does the inverse. The forward process is illustrated in Fig.~\ref{fig:NPFA}.

Given the input geometry, $\mathbf{G}_\text{R}$ and its paired neutral geometry $\mathbf{\bar G}_\text{R}$, our geometry encoder encodes it to the expression latent code via VAE sampling: $z_\text{R} = \mathcal{E}_\text{geo}(\mathbf{G}_\text{R})$. Then, the geometry decoder recovers the face geometry: $\mathbf{\tilde G}_\text{R}=\mathcal{G}_\text{geo}(\mathbf{\bar G}_\text{R}, z_\text{R})$. The training objective is simply a reconstruction loss: 
\begin{equation}
    \mathcal{L}_\text{recon,R}=\|\mathbf{\tilde G}_\text{R}-\mathbf{G}_\text{R}\|_2^2.
\end{equation}
Given a randomly sampled blendshape $w_\text{B}$ and neutral face $\mathbf{\bar G}_\text{B}$, we obtain the deformed expression $\mathbf{G}_\text{B}$ using personalized blendshapes.  Similar to the real data scenario, we extract the expression latent code ${\tilde z}_\text{B} =\mathcal{M}_\text{geo}(w_\text{B})$ and reconstruct the geometry through the geometry decoder, $\mathbf{\tilde G}_\text{B}=\mathcal{G}_\text{geo}(\mathbf{\bar G}_\text{B}, z_\text{B})$. We map the expression back by ${\tilde w}_\text{B} = \mathcal{M}'({\tilde z}_\text{B})$. The training objective is defined as:
\begin{equation}
    \mathcal{L}_\text{recon,B}=\|\mathbf{\tilde G}_\text{B}-\mathbf{G}_\text{B}\|_2^2 +\|{\tilde z}_\text{B}-z_\text{B}\|_2^2 + \|{\tilde w}_\text{B} -w_\text{B}\|_2^2.
\end{equation}

We use coordinate maps ~\cite{zhang2022video} to represent the geometries, which store the 3D position of each vertex on the 2D geometry map in the UV space. This representation can be converted to and from mesh representation using a fixed topology. Besides CG-compatible, it creates a more realistic and believable animation space than existing parametric facial models, due to its non-linearity and vertex-level granularity.

\begin{table}
  \caption{\textbf{4D datasets comparison}. Notice that DiffposeTalk~\cite{sun2023diffposetalk} is a combination of reconstructed TFHP~\cite{sun2023diffposetalk} and HDTF. EMOTE~\cite{EMOTE} is trained on reconstructed MEAD.}
  \label{tab:comparison_data}
  \resizebox{0.95\columnwidth}{!}{
    \begin{tabular}{lcccc}
    \toprule
              & Hours & Emotion & Head Pose & Language \\ 
    \midrule
    VOCASET &  0.5    &  \XSolidBrush    &     \XSolidBrush     &   EN            \\
    BIWI    &  1.7     &  \Checkmark     &   \Checkmark        &      EN         \\
    MultilFace & 2.8     &   \Checkmark    &    \Checkmark       &     EN        \\
    UUDaMM & 9.6   &   \XSolidBrush    &    \XSolidBrush       &     EN        \\
    DiffposeTalk  &  26.5  & \XSolidBrush  &  \Checkmark  & EN   \\
    EMOTE &  25.3   &  \Checkmark  &  \XSolidBrush  &  EN  \\
    \textbf{M2F-D (Ours)}  &   60.6      &    \Checkmark    &   \Checkmark    &     6         \\
    \bottomrule
    \end{tabular}
    }

\end{table}

\subsection{Image Facial Expression Extraction}

In addition to the geometry VAE, we train two vision encoders, $\mathcal{E}_\text{exp}$ and $\mathcal{E}_\text{pose}$, to extract unified expression latent code and head pose from RGB images. We freeze the geometry VAE and train the vision encoders under the supervision of both real images of RoM data and rendered images from geometries randomly generated by personalized blendshapes.

Specifically, given an image $\mathbf{I}_\text{R}$ in the RoM data with corresponding ground-truth geometry $\mathbf{G}_\text{R}$ and neutral $\mathbf{\bar G}_\text{R}$, we extract the expression latent code $\hat z_R=\mathcal{E}_\text{exp}(\mathbf{I}_\text{R})$, and head pose ${\hat p}_\text{R} = \mathcal{E}_\text{pose}(\mathbf{I}_\text{R})$. Then, we reconstruct the face geometry using our pretrained decoder, $\mathbf{\hat G}_\text{R} = \mathcal{G}_\text{geo}(\mathbf{\bar G}_\text{R}, {\hat z}_\text{R})$.  We utilize a differentiable renderer $\mathcal{R}$, to get the rendered image of the face: $\mathbf{\hat I}_\text{R} = \mathcal{R}(\mathbf{\hat G}_\text{R},{\hat p}_\text{R})$.  We define the training loss, $\mathcal{L}_\text{exp}$, as the combination of geometry loss and image loss:
\begin{equation}
    \mathcal{L}_\text{exp, R} = ~\|\mathbf{\hat G}_\text{R}-\mathbf{G}_\text{R}\|_2^2 + +\|\mathbf{\hat I}_\text{R}-\mathbf{I}_\text{R}\|_2^2.
\end{equation}
Similarly, with a randomly sampled geometry $\mathbf{G}_\text{B}$, we can extract its head pose ${\hat p}_\text{B} = \mathcal{E}_\text{pose}(\mathcal{R}(\mathbf{G}_\text{B},p_\text{B}))$, and expression code ${\hat z}_\text{B} = \mathcal{E}_\text{exp}(\mathcal{R}(\mathbf{G}_\text{B},p_\text{B}))$. We render the image by the same differentiable renderer $\mathbf{\hat I}_\text{B} = \mathcal{R}(\mathbf{\hat G}_\text{B},{\hat p}_\text{B})$. Our training objective is defined as:
\begin{equation}
    \mathcal{L}_\text{exp, B} = \|\mathbf{\hat G}_\text{B}-\mathbf{G}_\text{B}\|_2^2 \nonumber +\|\mathcal{R}(\mathbf{\hat G}_\text{B},{\hat p}_\text{B})-\mathcal{R}(\mathbf{G}_\text{B},p_\text{B})\|_2^2.
\end{equation}

After training, $\mathcal{E}_\text{exp}$ and $\mathcal{E}_\text{pose}$ can capture fine-grained expressions and head poses from in-the-wild videos, represented in the expression latent space, and map them to personalized expressions by $\mathcal{G}_\text{geo}$.
Owing to the rapid inference speed of GNPFA, we can efficiently extract high-quality and diverse expressions and head poses from in-the-wild videos.

\begin{figure*}[tbp]
  \centering
  \includegraphics[width=1.04\textwidth]{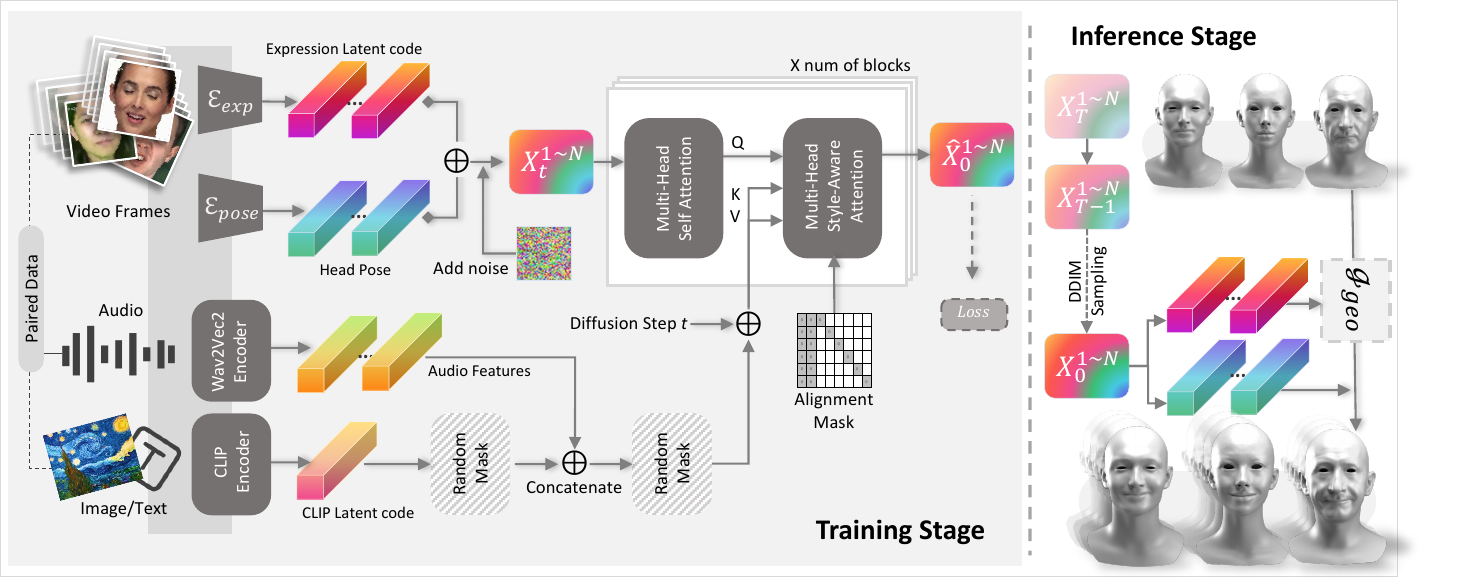}
  \captionof{figure}{\textbf{Architecture of Media2Face.} Our model takes audio features and CLIP latent code as conditions and denoise the noised sequence of expression latent code together with head pose i.e. head motion code. The conditions are randomly masked and subjected to cross-attention with the noisy head motion code. At inference, we sample head motion codes by DDIM. We feed the expression latent code to the GNPFA decoder to extract the expression geometry, combined with a model template to produce facial animation enhanced by head pose parameters.} \label{fig_overview}
\end{figure*}

\subsection{Latent-based Facial Animation Dataset}

We leverage a large collection of online video facial data with abundant audio and text labels, and use GNPFA to extract exact facial expressions and accurate head poses.  This allows us to avoid tedious annotations and thus easily augment the limited 4D facial animation dataset, which presents the Media2Face Dataset (M2F-D).

We retrieve both the expression latent codes and head poses from  MEAD ~\cite{kaisiyuan2020mead}, CREMA-D ~\cite{cao2014crema}, RAVDESS ~\cite{livingstone2018ryerson}, HDTF ~\cite{zhang2021flow} and Acappella ~\cite{montesinos2021cappella}. The MEAD dataset comprises talking-face videos of 60 actors and actresses expressing 8 different emotions at 3 varying intensity levels. The CREMA-D dataset comprises 7,442 distinct clips featuring 91 actors who delivered 12 sentences expressing 6 different emotions at 4 different intensity levels. The RAVDESS dataset consists of videos by 24 professional actors who vocalize speeches and songs, with a total of 7 and 5 emotions respectively.  The HDTF dataset is a collection of high-quality videos and the Acappella dataset encompasses solo singing videos.

To further increase the diversity of talking languages, we collect a 2-hour dataset from in-the-wild videos which contains 6 different languages: Chinese, French, German, Japanese, Russian, and Spanish. 
To allow for explicit head pose control under different scenarios, we capture a 1.6-hour dataset that consists of 14 speakers performing 14 different head movements including speaking, singing, nodding, shaking head, frowning, winking, etc.

Our M2F-D dataset has a total duration of over 60 hours at 30 fps. As shown in Table~\ref{tab:comparison_data}, and surpasses existing audio-visual datasets both in duration and diversity.

\section{Media2Face Methods}
\label{sec:method}

\textit{Media2Face} is a transformer-based latent diffusion model conditioning on multi-modal driving signals. 
It models the joint distribution of sequential head poses and facial expressions, i.e., full facial animation, and thus facilitates the natural synergy of poses and expressions. 
It also employs multi-conditioning guidance, enabling highly consistent co-speech facial animation synthesis with CLIP-guided stylization and image-based keyframe editing.

\subsection{Facial Animation Latent Diffusion Models}

As described in Fig.~\ref{fig_overview}, the expression latent code $\boldsymbol{z}^i_e$ and head pose $\theta^i$ are first extracted from each video frame $i$. Then, we concatenate them to form a single-frame facial animation state, denoted by $\boldsymbol{x^i}=[\boldsymbol{z}_e^i,\theta^i]$. The facial animation is thus formed by a sequence of states $\boldsymbol{X}^{1:N}=[x^i]_{i=1}^{N}$.

In the diffusion model, generation is modeled as a Markov denoising process. 
Here, $\boldsymbol{X}^{1:N}_{t}$ is obtained by adding noise to the ground truth head motion code $\boldsymbol{X}^{1:N}_0$ over $t$ steps. 
Our method models the distribution $p\left(\boldsymbol{X}^{1:N}_{0}|\boldsymbol{X}^{1:N}_{t}\right)$ to facilitate a stepwise denoising process. 
Similar to the approach in MDM ~\cite{tevet2022human}, we predict $\boldsymbol{X}^{1:N}_{0}$ directly. This prediction method allows us to introduce additional regularization terms to improve action consistency and smoothness. 

We employ large-scale pre-trained encoders to incorporate multi-modal conditions.
The raw speech audio is encoded by the pre-trained Wav2Vec2 ~\cite{baevski2020wav2vec} and aligned to the length of the facial animation sequence by linear interpolation, resulting in audio feature \(\boldsymbol{A}^{1:N}\). 
Besides, a text or an image serving as the prompt for talking style is encoded to CLIP latent code $\boldsymbol{P}$ by the pre-trained CLIP model ~\cite{radford2021learning}. 
Our Transformer-based denoiser learns to predict facial animation $\boldsymbol{X}^{1:N}_{0}$ conditioning on the concatenation of these multi-modal embeddings via the common style-aware cross-attention layers.
At each time step, the denoising process can be formulated as:
\begin{equation} \hat{\boldsymbol{X}}_0^{1:N}=\mathcal{G}\left(\boldsymbol{X}_t^{1:N},t,\boldsymbol{A}^{1:N},\boldsymbol{P}\right).
\end{equation}
To enable disentanglement of speech and prompt control, during training, two random masks are introduced for multi-conditioning classifier-free guidance ~\cite{ho2021classifier}. 
Initially, the CLIP latent code $\boldsymbol{P}$ undergoes the first random mask, which brings both stylized and non-stylized co-speech denoisers and enables style control disentangled with speech signals. 
Then, this masked code is concatenated with the audio feature $\boldsymbol{A}^{1:N}$. 
A second phase of random masking is applied to the final concatenated code, which similarly brings both speech-driven and non-speech-driven denoisers and facilitates adjusting speech content consistency strength.

\paragraph{Training}

We employ the simple loss ~\cite{ho2020denoising} as the main objective to train our models, which is defined as:
\begin{equation}
  \mathcal{L}_\text{simple}= \Vert \boldsymbol{X}^{1:N}_{0} - \boldsymbol{\hat{X}}^{1:N}_{0}\Vert ^2_2.
\end{equation}
Besides, we introduce a velocity loss ~\cite{Cudeiro_2019_CVPR} to enforce the model to produce natural transitions between adjacent frames, which is formulated as:
\begin{align}
  \mathcal{L}_\text{velocity}&=\left\Vert \left(\boldsymbol{X}^{2:N}_{0} - \boldsymbol{X}^{1:N-1}_{0}\right) - \left(\boldsymbol{\hat{X}}^{2:N}_{0} - \boldsymbol{\hat{X}}^{1:N-1}_{0}\right)\right\Vert ^2_2.
\end{align}
Furthermore, a smooth loss ~\cite{sun2023diffposetalk} is employed to enforce smoothness and reduce abrupt transitions and discontinuities:
\begin{align}
  \mathcal{L}_\text{smooth}&=\left\Vert\boldsymbol{\hat{X}}^{3:N}_{0} + \boldsymbol{\hat{X}}^{1:N-2}_{0} - \boldsymbol{\hat{X}}^{2:N-1}_{0} \right\Vert ^2_2.
\end{align}
Overall, the denoiser is trained with the following objective:
\begin{align}
      \mathcal{L} = \lambda_\text{simple}\mathcal{L}_\text{simple} + \lambda_\text{velocity}\mathcal{L}_\text{velocity} + \lambda_\text{smooth}\mathcal{L}_\text{smooth},
\end{align}
where $\lambda_\text{simple}$, $\lambda_\text{velocity}$, and $\lambda_\text{smooth}$ are hyper-parameters serving as loss weights to balance the contributions from these terms.

\paragraph{Inference}

During the denoising process, our model combines two types of guidance, the main speech audio and the additional text/image style guidance, with the classifier-free guidance technique ~\cite{ho2021classifier}:
\begin{align}
      \hat{\boldsymbol{X}}_0^{1:N} &= 
       (1-\textbf{s}_{A}-\textbf{s}_{P})\cdot\mathcal{G}\left(\boldsymbol{X}_t^{1:N},t\right)  \nonumber \\ 
      &+  \textbf{s}_{A}\cdot\mathcal{G}\left(\boldsymbol{X}_t^{1:N},t,\boldsymbol{A}^{1:N}\right)
      + \textbf{s}_{P}\cdot\mathcal{G}\left(\boldsymbol{X}_t^{1:N},t,\boldsymbol{A}^{1:N},\boldsymbol{P}\right),
\end{align}
where $\textbf{s}_{A}$ and $\textbf{s}_{P}$ are two strength factors to adjust speech and style guidance strength respectively.
The last two terms provide both non-stylized and stylized predictions within the same speech inputs, which implies the disentangled style control beyond the speech content.

\paragraph{Overlapped batching denoising}

To reduce the inference time for real-time applications, we employ \textit{batching denoising}, a technique akin to the batching denoising step introduced in StreamDiffusion ~\cite{kodaira2023streamdiffusion}, and further extend it to \textit{overlapped batching denoising}, to process exceedingly long audios in a single denoising pass, by segmenting audio into overlapped windows.
The overlapped batching denoising approach transforms the traditionally multiple, autoregressive sequence generation tasks into a parallelizable endeavor. Within the confines of VRAM capacity, its processing time does not increase linearly with the length of the audio, thereby significantly enhancing the speed of head motion generation.

\subsection{Conditional Facial Animation Editing}

\begin{figure}
  \includegraphics[width=0.95\columnwidth]{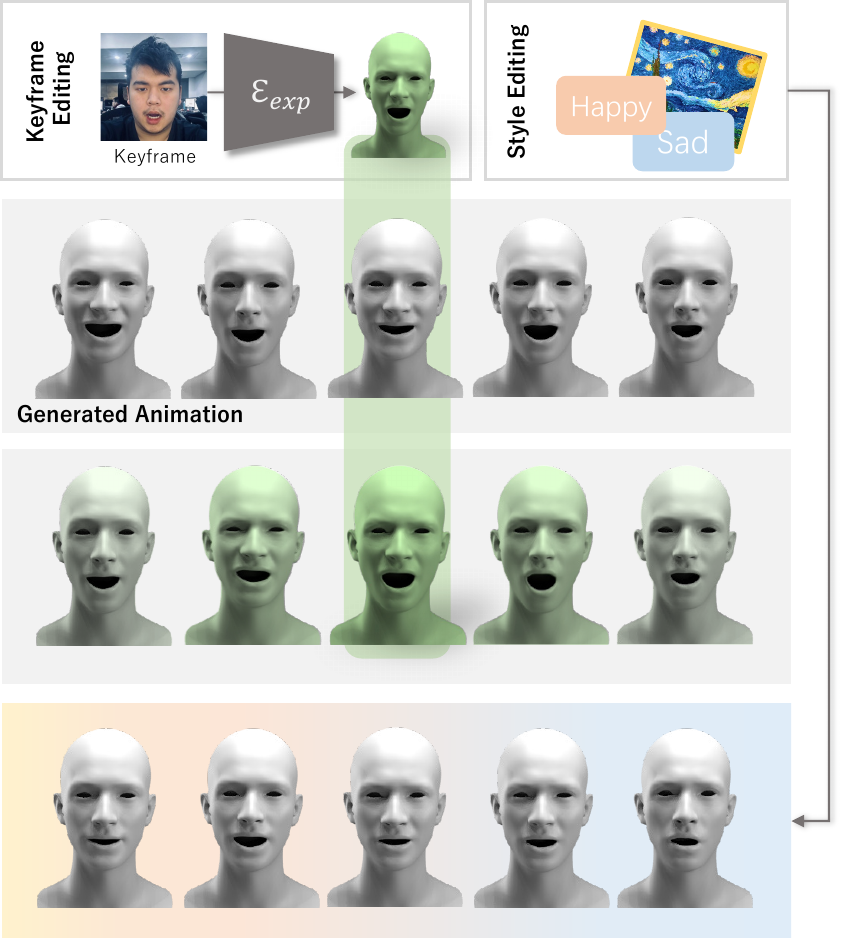}
  \caption{\textbf{Application show case.} We can fine-tune the generated facial animation (\textit{Row 2}) by 1. extracting key-frame expression latent codes through our expression encoder (\textit{Row 3}), 2. providing per-frame style prompts through CLIP (\textit{Row 4}, \textit{Left}: happy, \textit{Right:} Sad). The intensity and range of control can be adjusted using diffusion in-betweening techniques.}
  \label{fig:application}
\end{figure}

Media2Face achieves fine-grained control of generation through keyframe editing and text/image guidance. As illustrated in Fig. ~\ref{fig:application}, we use GNPFA and CLIP to extract the conditions from face images and text/image prompts and leverage classifier-free guidance to control the diffusion process.

\paragraph{Keyframe editing}
We can modify the keyframes of the generated animation and smoothly integrate them with the corresponding lip movements using diffusion inpainting technique \cite{Lugmayr_2022_CVPR} in the temporal domain. Fig.~\ref{fig:application} illustrates an example of modifying a keyframe retrieved from an image using GNPFA. 
Similarly, this ability can be generalized as \textit{sequential composition} in \cite{shafir2023human} to diffuse animations from different sources together. Please refer to our supplementary video.

\paragraph{CLIP-guided style editing}

Utilizing an in-betweening technique in~\cite{tevet2022human,ao2023gesturediffuclip}, our approach enables the application of diverse style controls across different frames within an audio segment.  By assigning distinct style prompts to individual frames and employing a gradient mask during each diffusion step, we seamlessly and naturally integrate the sampling results of various prompts. This methodology ensures a coherent transition of style influences throughout the audio sequence.

\section{Results}
\label{sec:experiments}
\begin{figure*}
  \includegraphics[width=\textwidth]{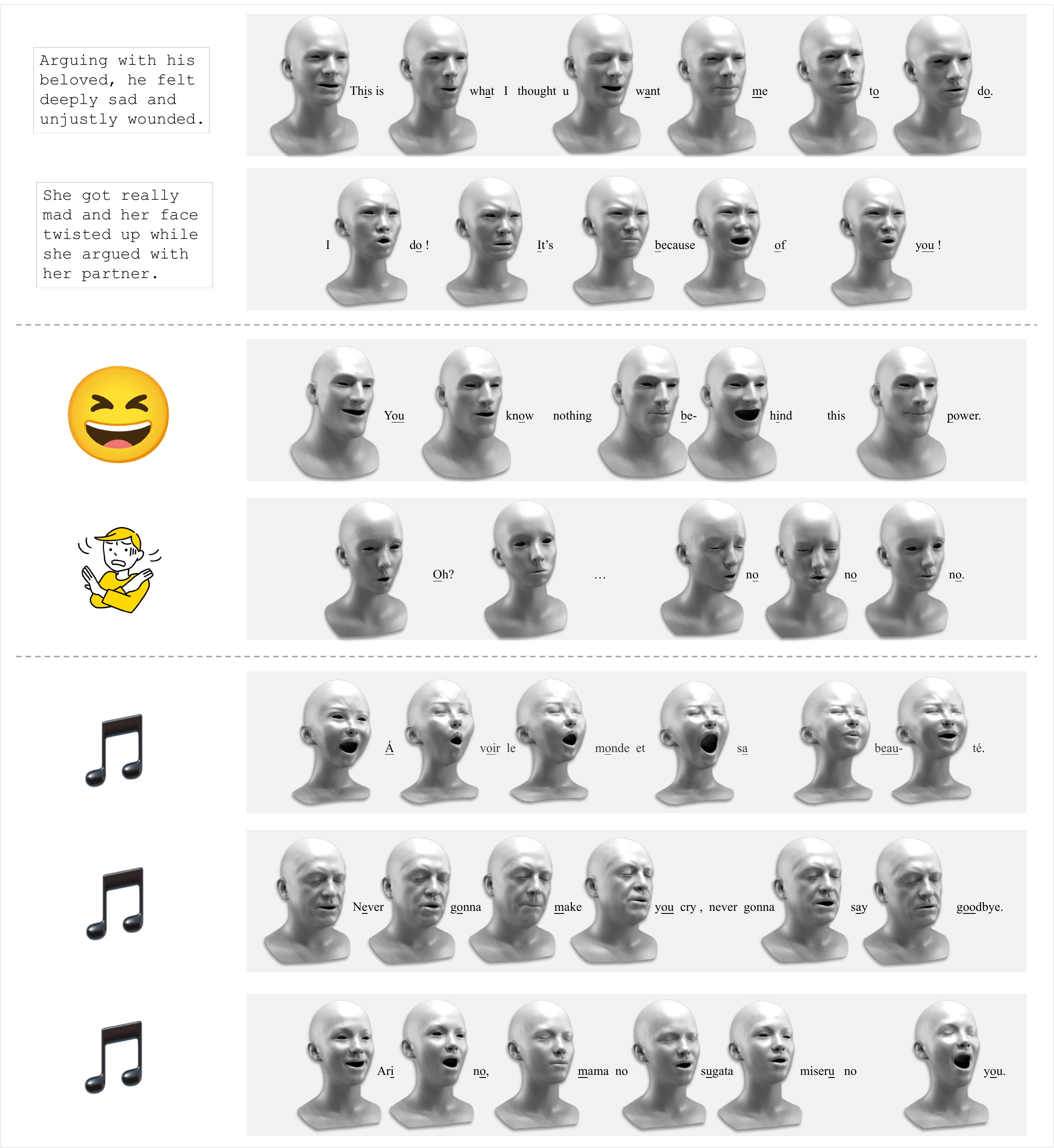}
  \label{fig:results_gallery}
  \caption{\textbf{Result gallery.} We generate vivid dialogue scenes (\textit{Row 1,2}) through scripted textual descriptions. We synthesize stylized facial animations (\textit{Row 3,4}) through image prompts, which can be emoji or even more abstract images. We also perform emotional singing in France, English and Japanese(\textit{Row 5-7}). For more results, please refer to the supplementary video.}
  \label{fig:gallery}
\end{figure*}

In Fig.~\ref{fig:gallery}, we showcase multiple audio-driven animations as well as style-based generation results based on text and image prompts. Please refer to the supplementary video for more results.

\subsection{Implementation Details}

For GNPFA, we follow Dreamface ~\cite{zhang2023dreamface} to design our geometry VAE, vision encoders share the same architecture as the geometry encoder. Training of the geometry VAE and vision encoders take 10 days and 96 hours to converge respectively on Nvidia A6000 GPU, using an AdaBelief optimizer. GNPFA can inference on Nvidia RTX 3090 GPU at about 500 fps.
For Media2Face, we employ an eight-layer transformer decoder for the denoiser, utilizing four attention heads. The feature dimension is 512, and the window size is $N=200$ at 30 fps. During training, our models follow a cosine noise schedule with 500 noising steps.  Media2Face is trained on Nvidia RTX 3090 GPU for 36 hours using an AdamW optimizer. We set $\lambda_\text{smooth} = 0.01, \lambda_\text{velocity} = 1, \lambda_\text{simple}=1$. During inference time, we set $\textbf{s}_{A}=2.5$, $\textbf{s}_{P}=1.5$. We achieved over 300 fps offline and 30fps in real-time on Nvidia RTX 3090 GPU.

\subsection{Comparisons}

We compare Media2Face with several state-of-the-art facial animation methods. We separate a two-hour segment from the M2F-D as the test set and keep it with a similar data structure as the training set. For 3D methods, we compared with FaceFormer ~\cite{faceformer2022}, CodeTalker ~\cite{xing2023codetalker}, FaceDiffuser ~\cite{FaceDiffuser_Stan_MIG2023} and EmoTalk ~\cite{peng2023emotalk} and use their pre-trained model as the baseline. We unify all results to the same FLAME topology for fair comparisons. We also compared the quality of generated head poses with SadTalker ~\cite{zhang2022sadtalker}, a 2D talking face generation method that incorporates head movements.

\subsubsection{Quantitative Comparisons}

To measure lip synchronization, we employ Lip Vertex Error (LVE) ~\cite{richard2021meshtalk}, calculating the maximum L2 error across all lip vertices for each frame. 
Upper Face Dynamics Deviation (FDD)~\cite{xing2023codetalker} measures the diversity of expressions by comparing the standard deviation of each upper face motion over time between the synthesized and the ground truth.
To evaluate the synchronization between audio and generated head pose, we utilize Beat Alignment (BA)~\cite{sun2023diffposetalk,zhang2022sadtalker}, which computes the synchronization of head pose beats between the generated and the ground truth.
As shown in Table~\ref{tab:comparison-quantitative}, our method surpasses existing methods in terms of lip accuracy, facial expression stylization, and the synthesis of rhythmic head movements.

\begin{table}
  \small
  \caption{\textbf{Quantitative comparisons and evaluations}. Notice that the BA metric is not utilized for FaceFormer, CodeTalker, FaceDiffuser, and EmoTalk, as they do not generate head poses. Also, metrics related to vertices are not utilized for SadTalker due to its different facial topology.}
  \label{tab:comparison-quantitative}
  \resizebox{0.95\columnwidth}{!}{
    \begin{tabular}{cccccc}
    \toprule
    Methods        & LVE(mm)$\downarrow$ & FDD($\times10^{-5}$m)$\downarrow$ & BA $\uparrow$ \\
    \toprule
    FaceFormer     &  18.19   &  21.37   &  N/A  \\
    CodeTalker     &  16.74   &  21.95   &  N/A  \\
    FaceDiffuser   &  16.33   &  22.38   &  N/A  \\
    EmoTalk        &  14.61   &  17.84   &  N/A  \\
    SadTalker      &  ---     &   ---    & 0.219 \\
    \midrule
    Ours w/o CFG   & 10.67  &  16.69   & 0.166  \\
    Ours w/o GNPFA & 14.89  &  12.81   &  0.198  \\
    \midrule
    Ours w/ 10\% data &  10.75   &  20.65   &  0.170  \\
    Ours w/ 40\% data &  10.55   &  18.32   &  0.208  \\
    Ours w/ 70\% data &  \textbf{10.43}   &  14.98   &  0.221  \\
    \midrule
    \textbf{Ours}  &  10.44   &  \textbf{12.21}   &  \textbf{0.254}   \\
    \bottomrule
    \end{tabular}
    }
\end{table}

\subsubsection{Qualitative Comparisons}
\begin{figure}[H]
  \includegraphics[width=0.95\columnwidth]{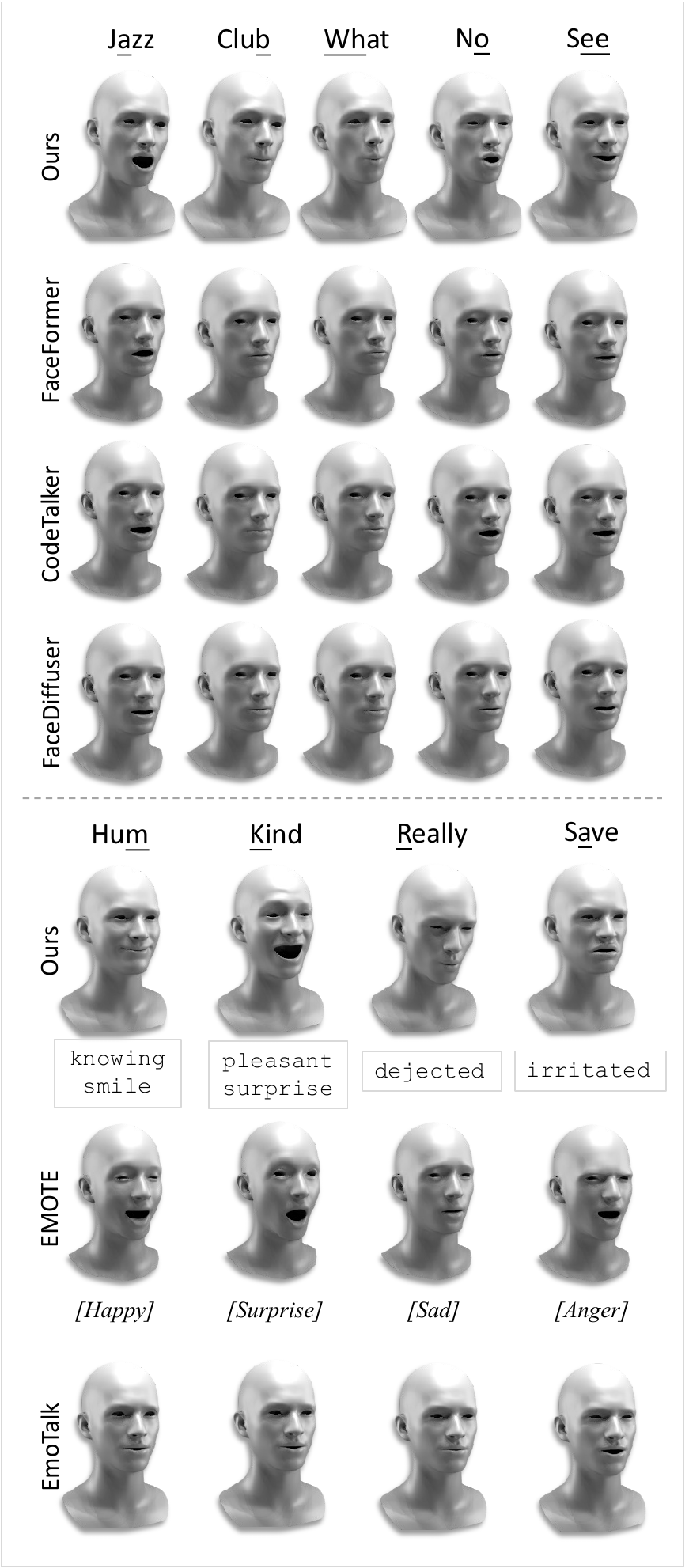}
  \caption{\textbf{Qualitative comparison.} \textit{Top:} Comparing with emotion-blind methods, we use a neutral prompt to feed to Media2Face. \textit{Bottom:} Comparing with emotion-aware methods. We utilize text prompts for Media2Face and assign corresponding emotion labels to EMOTE. Notice that EmoTalk extracts emotional features from audio that cannot be manually assigned.}
  \label{fig:comparison_qualitative}
\end{figure}

We show our qualitative comparison in Fig.~\ref{fig:comparison_qualitative}. Compared with emotion-blind methods, (FaceFormer, CodeTalker, FaceDiffuser), our method can generate not only more accurate lip movement but also micro-expressions under neutral conditions (eye blink, eyebrow gesture). 
Compared with emotion-aware methods, (EMOTE, EmoTalk), our method demonstrates a more vibrant and natural expression of emotions and facial details while maintaining lip shape accuracy. Notice that our method also generates head poses highly synchronized with the given conditions(raise the head in surprise and lower it in sadness).

\subsection{Ablation study}
 We conduct following ablation experiments to evaluate our key components:(1) \textit{Ours w/o GNPFA}: We train Media2Face on linear blendshapes obtained from the GNPFA mapping network $\mathcal{M}'$. (2) \textit{Ours w/o CFG}: We inference Media2Face without classifier-free guidance. As shown in Table ~\ref{tab:comparison-quantitative}, the removal of GNPFA leads to a significant degradation in LVE, validating the effectiveness of GNPFA on modeling accurate lip shape. Inference without CFG has bad performance on FDD since the model fails to generate stylized head motions. 
We also train Media2Face on 10\%, 40\%, 70\% of M2F-D.  As shown in Table~\ref{tab:comparison-quantitative}, model performance on FDD and BA increases during dataset scaling up while that on LVE remains steady. This validates our hypothesis that while the model can learn precise lip-sync animation on small datasets, it requires learning from a large amount of rich-conditioned data to generate animation with realistic expressions, diverse emotions, and appropriate head movements.

\subsection{User Study}

We conduct 30 diverse audio samples, including dialogues, speeches, and songs and invert 100 participants. We ensure fair comparisons by employing the same shader and template for all generated geometries. Participants ensure side-by-side animations with other methods, assessing Media2Face with three conditions: with a specific style prompt for each audio, with a neutral prompt, and without any prompts and head pose animation. Our model demonstrates superior preference ratings: over 90\% for general cases, 80\% without specific style prompts, and 70\% in the absence of specific style prompts and head pose, underscoring the effectiveness of head pose generation and our style prompt.

\section{Conclusions}
\label{sec:conclusions}

In this paper, we present Media2Face, pushing the boundary of diffusion models for realistic co-speech facial animation synthesis with rich multi-modal conditionings. 

To enhance the diffusion models with high-quality facial animation data, we introduce GNPFA, a facial VAE with a latent neural representation of facial expressions and head poses, pre-trained on a wide array of facial scanning data. GNPFA is then utilized to extract high-quality expressions and head poses from a mass of accessible facial videos from various resources.
It brings M2F-D, a large, diverse, and scan-level 3D facial animation dataset with abundant speech, emotion, and style annotations, over 60 hours. 
Finally, we train our Media2Face model in GNPFA latent space with M2F-D dataset.  
Media2Face integrates diverse media inputs as conditions including audio, text, and image, which flexibly control facial emotion and style while preserving high-quality lip-sync with speech. 
The experimental results demonstrate the effectiveness of Media2Face and showcase various related applications, such as reconstructing dialogue situations and multi-modality conditional editing.
We believe Media2Face is a significant step towards realizing realistic human-centric AI virtual companions with strong emotional connection and resonance with us humans.

\clearpage

\begin{figure}
  \includegraphics[width=\columnwidth]{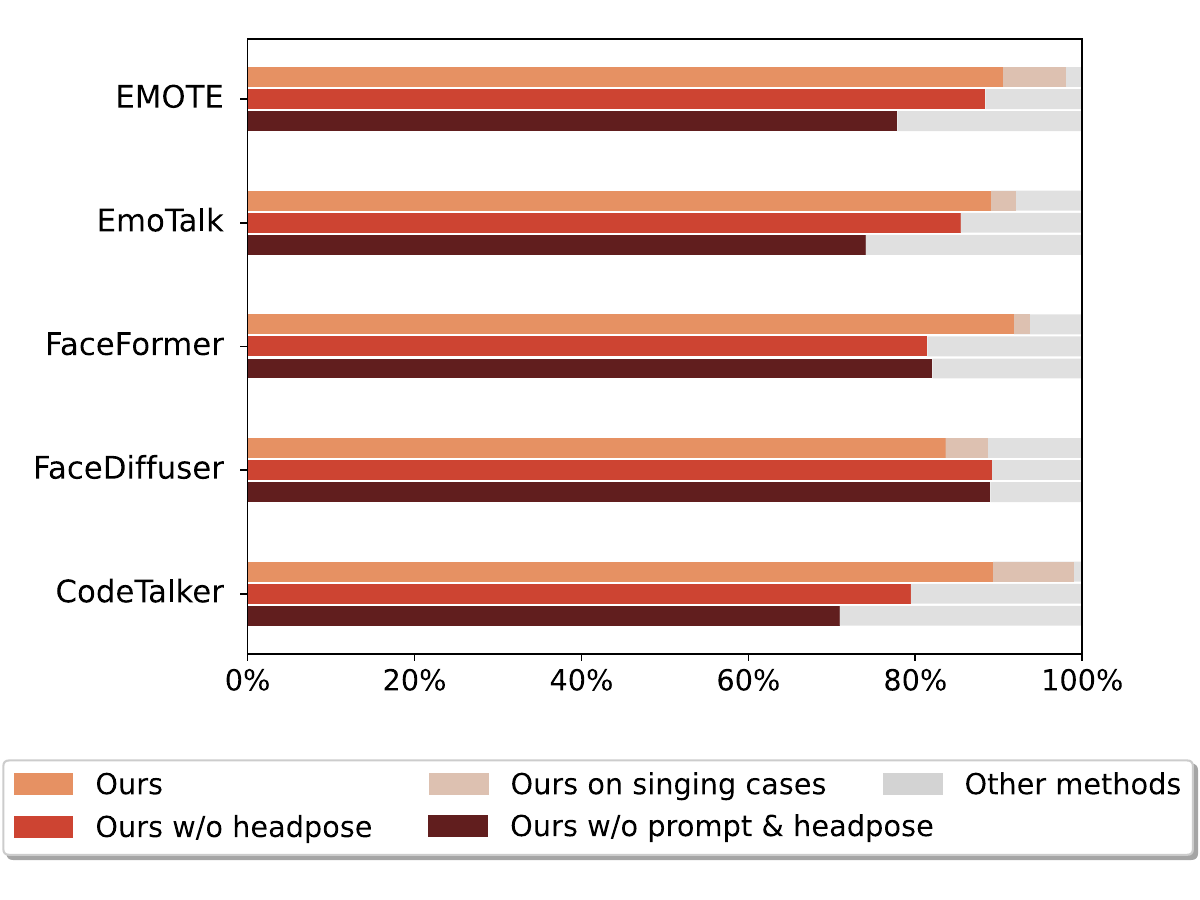}
  \caption{\textbf{User study result.} Note how our method has demonstrated overwhelming superiority in the singing cases, showcasing the model's ability to generate rich emotions and rhythmic head movements. }

  \label{fig:user-study}
\end{figure}

\begin{figure}
  \includegraphics[width=0.95\columnwidth]{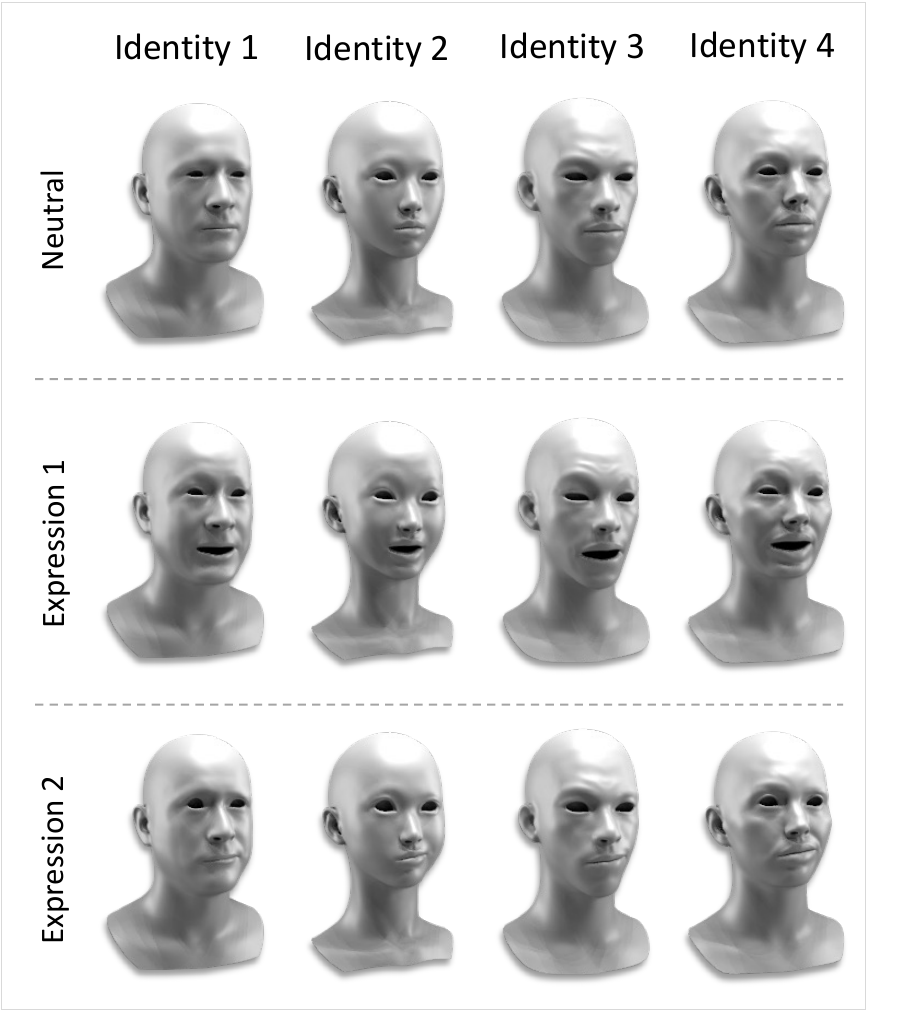}
  \caption{\textbf{Retargeting to various identities.} Thanks to GNPFA, we can further generate personalized and nuanced facial mesh, which can fit various identities across different genders, ages, and ethnicities. Note the differences in facial details among different identities, notably the different wrinkles.}
  \label{fig:retargeting}
\end{figure}

\bibliographystyle{ieee_fullnames}
\bibliography{main}

\end{document}